\documentclass{article}
\usepackage{spconf,amsmath,graphicx}

\usepackage{amssymb}
\usepackage{comment}
\usepackage{xcolor,booktabs}
\usepackage{multirow}
\usepackage[shortlabels]{enumitem}
\usepackage{hyperref}

\hypersetup{
    colorlinks=true,
    linkcolor=blue,
    filecolor=magenta,      
    urlcolor=cyan,
    pdftitle={Overleaf Example},
    pdfpagemode=FullScreen,
    }
    
\usepackage{bbm}

\newif\ifcomments
\commentstrue
\ifcomments
\providecommand{\jb}[1]{{\protect\color{red}{[JB: #1]}}}

\providecommand{\mi}[1]{{\protect\color{blue}{[MI: #1]}}}
\providecommand{\yb}[1]{{\protect\color{purple}{[YB: #1]}}}
\else
    \providecommand{\jb}[1]{}
    \providecommand{\mi}[1]{}
    \providecommand{\yb}[1]{}
\fi

\newcommand\commentout[1]{}

\title{Scene Graph to Image Generation with \\ Contextualized Object Layout Refinement}
\name{Maor Ivgi$^*$\thanks{$^*$ equal contribution}, Yaniv Benny$^*$, Avichai Ben-David, Jonathan Berant, Lior Wolf}
\address{Blavatnik School of Computer Science, Tel-Aviv University}

\begin{document}
%
\newcommand\cocostuff{\textsc{COCO-Stuff} }

\maketitle
\begin{abstract}
Generating images from scene graphs is a challenging task that attracted substantial interest recently. Prior works have approached this task by generating an intermediate layout description of the target image. However, the representation of each object in the layout was generated independently, which resulted in high overlap, low coverage, and an overall blurry layout. We propose a novel method that alleviates these issues by generating the entire layout description gradually to improve inter-object dependency.
We empirically show on the \cocostuff dataset that our approach improves the quality of both the intermediate layout and the final image. Our approach improves the layout coverage by almost 20 points, and drops object overlap to negligible amounts. 
\end{abstract}
\section{Introduction}\label{sec:intro}

Synthesizing images from natural language descriptions
has received substantial attention recently \cite{zhang2017stackgan, zhang2017stackgan++, li2019object, xu2018attngan},
as it has wide applicability for content generation. However, it has been shown that models that accept textual descriptions as their input fail to produce images with multiple detailed objects with complex relations \cite{zhang2017stackgan, zhang2017stackgan++, xu2018attngan}. Thus, \emph{scene graphs}~\cite{johnson2015image}, i.e. graphs where nodes correspond to entities and edges describe relations between them, were proposed~\cite{Johnson_2018_CVPR} as an intermediate representation of the desired image. This approach has been widely adopted \cite{ashual2019specifying, Herzig2020LearningCR, Tripathi2019UsingSG} for this task.

When generating images from scene graphs (SG), there are three main desiderata: (i) \emph{Photo-realism}: the image should look natural with salient objects, (ii) \emph{Correctness}: the image should contain the objects and relations specified in the SG, and (iii) \emph{Diversity}: because an SG is an underspecified representation compatible with many output images, a model should reflect that in its output distribution. Current models for SG-to-image generation invariably combine a supervised learning objective at training time. Specifically, given an SG and an image they predict for each object separately the \emph{exact} location and shape from the gold semantic layout to produce the ground truth image. 
Although this can achieve correctness for simple geometric relations, it inevitably results in poor quality image-layout due to the under-specificity of the SG. In particular, many distinct images can be represented by the same SG, thus maximum likelihood based techniques result in a blurry average of object shapes and positions across possible images. Such generations are likely to exhibit low resolution, low coverage, and high inter-object overlap.
Moreover, due to the strict specification of the prediction task, true diversity is inherently impossible.

In this work we propose two main technical contributions to solve these issues: (i) To address the diversity issue, we reduce the dependence on supervised losses and shift towards adversarial ones.
In particular, rather than predicting the box and mask of each object according to the target image, we use an adversarial network as a discriminator. It ensures that the generated object layout is truthful to the required object class in both position and shape and that the relation between every pair of objects is sensible and obeys the constraints dictated by the SG. (ii) To address the quality issue, we introduce a novel method to perform high-resolution layout generation.
It incorporates the ability of Graph Convolution Networks (GCNs)~\cite{kipf2016semi} to work on variable-shaped structured graphs and contextualizes the state of all objects with CNN-based generators. Using this layout refinement network, we fuse predicted object layouts such that each remains true to its class and respects its dictated relations, while maintaining high coverage and few overlaps. We stack multiple copies of this block and present \textit{Contextualized Objects Layout Refiner} (COLoR): the first model to generate layouts directly from SGs without any intermediate steps such as boxes and masks.

\begin{figure*}[t]
\includegraphics[width=\linewidth, trim={25 700 25 25}, clip]{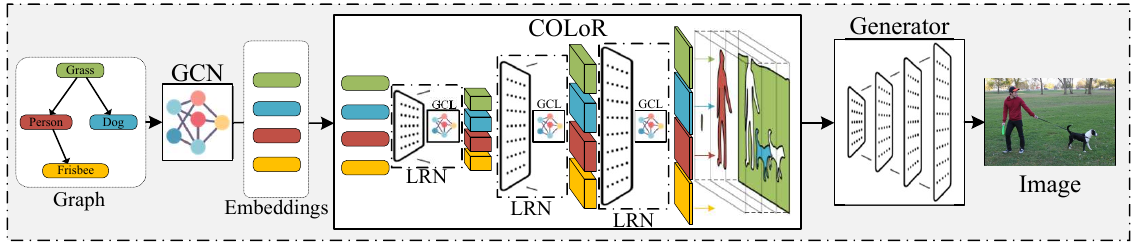}
\vskip -0.1in
\caption{Our Scene Graph to Image architecture. A GCN encodes the SG nodes into individual embeddings. The COLoR module upsampled the embeddings with intermediate GC layers into the scene layout. The SPADE generator then produces the image.}
\label{fig:architecture}
\vskip -0.2in
\end{figure*}

\section{Background}\label{sec:background}
We now describe the architecture of prior work SG2Im~\cite{Johnson_2018_CVPR}, which we build upon, and formally define the task. 

\noindent{\bf Problem setup\quad}
Our goal is to train a model $p_\theta(I \mid G, s)$ that takes as input an SG and a random seed $s$ and outputs an image $I$. Given a vocabulary $\mathcal{C}$ of object categories and a set $\mathcal{R}$ of possible relations, an SG is a directed graph $G=(\mathcal{O},\mathcal{E})$, where each node $o \in \mathcal{O}$ is an object associated with a class by $C: \mathcal{O} \rightarrow \mathcal{C}$, and edges $\mathcal{E} \subseteq \mathcal{O} \times \mathcal{O}$ represent directed relations between objects, associated with a type through $R: \mathcal{E} \rightarrow \mathcal{R}$.

During training, the available information for every Image $I$ is a segmentation mask, identifying each pixel in the image to a unique object and its class. This can be used to generate multiple SGs for each image by computing geometric relations between objects and randomly sampling from all possible edges in the complete graph. In addition, this segmentation mask is used to infer the layouts, masks, and bounding boxes for all objects in the image, as defined below.

\noindent{\bf Layout generation\quad}
An \textit{Image Layout} is a mapping of each pixel in the image to a specific object. Given an object in the layout, we define an \textit{object layout} $\textbf{l} \in [0,1]^{H \times W}$ as a mapping over the image, indicating pixels that belong to the object. Higher values signify stronger presence of the object. Ideally (as is the case in the annotated layouts), $\textbf{l}$ is binary. We then define the \textit{Object Box} $\textbf{b} \in [0,1]^{4}$ as the minimal axis-aligned bounding box in relative coordinates of all active pixels in $\textbf{l}$. Finally, cropping $\textbf{l}$ using $\textbf{b}$ and projecting it into $[0,1]^{W_m \times W_m}$ where $W_m < \min(H,W)$, we get the \textit{Object Mask} $\textbf{m} \in [0,1]^{W_m \times W_m}$, which describes its shape, with higher-value pixels corresponding to the existence of the object in said pixel. We note that though $\textbf{b}$ and $\textbf{m}$ are derived uniquely from $\textbf{l}$, it is possible to approximate $\textbf{l}$ by performing the inverse projection of $\textbf{m}$ into $[0,1]^{H \times W}$ according to $\textbf{b}$.

\noindent{\bf Scene Graph to Image\quad}
\label{subsec:sgtoimage}
We build on SG2Im~\cite{Johnson_2018_CVPR}, which includes the following steps: (a) The SG is augmented with an additional dummy node which is connected to all other nodes through an outgoing dummy relation to ensure graph connectivity. (b) Every node and edge in the SG is replaced by a learned embedding $v \in \mathbb{R}^d$ based on its class.
(c) The graph is fed to a GCN, which produces a new embedding $\Tilde{v}$ for each object (node) in the SG. (d) The embeddings $\Tilde{v}_1, \dots, \Tilde{v}_n$ are fed into the layout predictor consisting of two separate decoders, one predicts a bounding box location $\hat{b}_i$, and another predicts a mask $\hat{m}_i$. Those are used to compute $\hat{l}_i$ as explained above. The embedding $\tilde{v}_i$ is multiplied element-wise with $\hat{l}_i$ producing $\hat{\ell}_i \in [0,1]^{H \times W \times d}$.
(e) The extended layouts $\hat{\ell}_1, \dots, \hat{\ell}_n$ are summed element-wise to produce a coarse image layout $\hat{\textbf{l}} \in [0,1]^{H \times W \times d}$.
(f) $\hat{\textbf{l}}$ is fed along with $z$ random noise channels into a Cascaded Refinement Network (CRN)~\cite{chen2017photographic}, predicting the final image $\hat{I}$.

In \cite{Johnson_2018_CVPR}, the model is trained with six loss functions: three adversarial loss functions that evaluate object realism, the ability to correctly classify objects, and image similarity to real images. The other three use strong supervision and force the model to predict boxes and masks which are similar to those in the ground truth image $\textbf{I}$. Those are:
\setlength{\abovedisplayskip}{3pt}
\setlength{\belowdisplayskip}{3pt}
\begin{equation}
\begin{split}
\mathcal{L}_{\text{box}} &= \sum\nolimits_{i=1}^{n}{\|\textbf{b}_i - \hat{b}_i\|_2}, \quad \mathcal{L}_{\text{pix}} = \|\textbf{I} - \hat{I}\|_1 \\
\mathcal{L}_{\text{mask}} &= \sum\nolimits_{i=1}^{n}{\sum\nolimits_{h,w}^{W_m,W_m}{\text{BCE}(\textbf{m}_{i,h,w}, \hat{m}_{i,h,w})}}
\end{split}
\end{equation}
where $\hat{b}_i, \hat{m}_i, \hat{I}$ are the predicted boxes, masks and Image and $\text{BCE}(p, \hat{p})=-p\log(\hat{p})-(1-p)\log(1-\hat{p})$.

\section{Method}
\label{sec:method}

\begin{figure}[t]
\begin{center}
\centerline{\includegraphics[width=\linewidth]{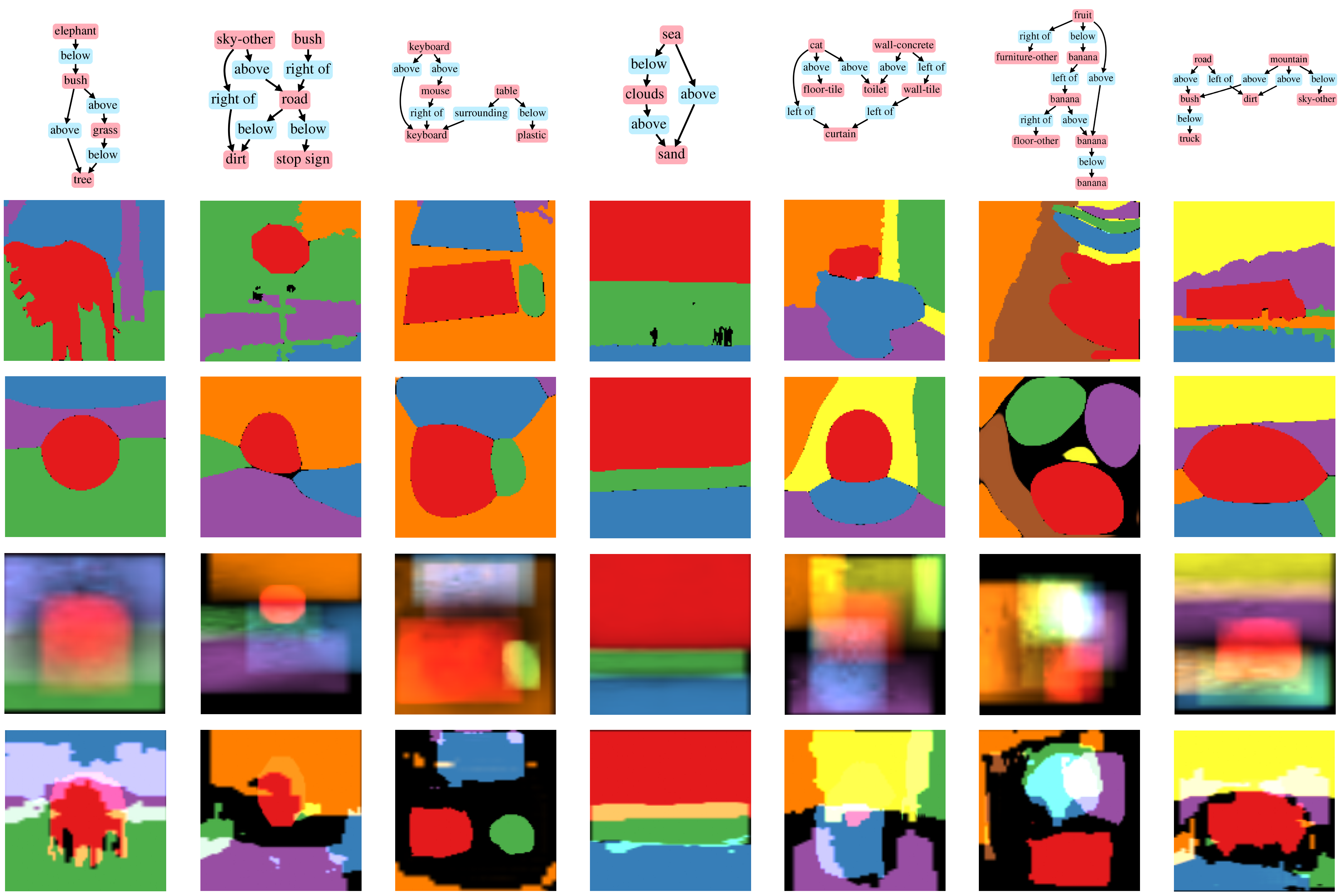}}
\vskip -0.1in
\caption{Comparison of layout generation. Layout values are in $[0,1]$, depicted as color opacity, thus making overlaps visible. Unassigned pixels are black. Rows from top to bottom: Input Scene graph, true layout, COLoR, SG2Im \cite{Johnson_2018_CVPR}, Grid2Im \cite{ashual2019specifying}.}
\label{fig:supervised-models-predictions}
\end{center}
\vskip -0.4in
\end{figure}

One major drawback of using the aforementioned supervised loss functions is the underlying assumption that for every SG there exists (in the dataset) \textbf{at most one} corresponding image layout. However, in \cocostuff~\cite{caesar2018coco} which is commonly used for this task, this is far from true, as $73\%$ of the images contain a (multi) set of objects that is shared with many other images in the data and may result in identical SGs. Further, over $25\%$ of the SGs match multiple different layouts, and almost 10\% of the SGs describe over 10 different layouts. Thus, a model that maximizes the likelihood of a layout given an SG will be pushed towards predicting the mean of the bounding boxes in the layouts that occur in the training data, and similarly, the average mask. Because the location and shapes of layout substantially vary across images, the model eventually will ignore the context and predict a general location for each object with no distinct shape as can be seen in Figure.~\ref{fig:supervised-models-predictions}.

To overcome this difficulty, we remove most of the loss terms that are applied with respect to the \textbf{exact} ground truth layout used (\S\ref{subsec:sgtoimage}) and reduce the weight of the rest. Instead, we add adversarial loss functions that encourages the model to generate photo-realistic images that respect the original SG, without forcing it to learn a single SG2Im mapping. The proposed discriminator are applied on the predicted object layouts $\hat{{l}}_i$. We find that some strong supervision is beneficial to cope with issues that are linked to cold-start.

\noindent{\bf Pairwise Layout Discriminator\quad}
The main source of training signal in our method is an adversarial network which teaches the generator to be spatially aware, creating objects without overlap that respect the relations set by the SGs. It follows the AC-GAN~\cite{odena2017conditional} adversarial loss pattern. Given a pair of neighboring objects in the SG, the discriminator accepts their object layouts $\textbf{l}_i, \textbf{l}_j \in [0,1]^{H \times W}$ and their class labels $c_i, c_j \in \mathcal{C}$. It predicts whether this pair comes from a real or a generated layout and classify the relation between the two. Since low-quality layouts will be easily recognized as fake, it also improves the quality of all object layouts individually. 
In particular, The discriminator $D_{l}$ performs a mapping $D_{l} : ([0,1]^{H \times W}, \{0,1\}^{|\mathcal{C}|})^2 \rightarrow [0,1] \times [0,1]^{|\mathcal{R}|}$.
Let $(\hat{y}, \hat{r}) = D_{r}((\textbf{l}_i, c_i), (\textbf{l}_j, c_j))$ be the prediction of the discriminator (real vs. fake and relation prediction) on its input. Let $r\in \{0,1\}^{|\mathcal{R}|}$ be the true relation and $y = \mathbbm{1}_{real}$.
\begin{equation}
\begin{split}
    \mathcal{L}_{D_l}^{d} &= \text{BCE}(y, \hat{y}) + \text{CE}(r, \hat{r}) \\
    \mathcal{L}_{D_l}^{g} &= \text{BCE}(1, \hat{y}) + \text{CE}(r, \hat{r})
\end{split}
\label{eq:D_l}
\end{equation}
where the discriminator trains on real and fake pairs, and the generator minimizes the loss over generated pairs only.

\noindent{\bf Losses\quad}
To complement our discriminator and encourage the model to assign objects to every pixel of the image and refrain from overlap, we introduce the \textit{Layout Coverage Regularization} $\mathcal{L}_{\text{reg}} = \mathcal{L}_{\text{coverage}} +\lambda \cdot \mathcal{L}_{\text{overlap}}$. Given $\hat{l}_1,\dots,\hat{l}_n \in [0,1]^{H \times W}$ we define the summed image layout $\hat{L}=\sum_{i}^{n}{\hat{l_i}} \in [0,n]^{H \times W}$ which gives the following definitions:
\begin{align}
\mathcal{L}_{\text{coverage}} &= \sum\nolimits_{h,w}^{H,W} { \mathbbm{1}[\hat{L}_{h,w} \le 1] \cdot \left ( 1- \hat{L}_{h,n} \right ) } \\
\mathcal{L}_{\text{overlap}} &= \sum\nolimits_{h,w}^{H,W} { \mathbbm{1}[\hat{L}_{h,w} > 1] \cdot \left ( \hat{L}_{h,n} - 1 \right ) }
\label{eq:layout}
\end{align}
The loss reaches 0 if the layouts weights in every pixel sum to exactly 1, and grows as the coverage drops or overlap increases. Since $\mathcal{L}_{\text{overlap}}$ is unbounded and often grows larger than $\mathcal{L}_{\text{coverage}}$, we suppress its contribution by setting $\lambda=0.4$ which we found achieves the best tradeoff between the terms. 

In addition, we found that when tasked to generate layouts with only the losses in equations \eqref{eq:D_l} and \eqref{eq:layout}, the generator fails to learn how to create coherent layouts, and the discriminator falls back to classify fake layouts based on spurious artifacts in the layouts. We attribute this issue to cold-start problem, and mitigate it by adding small weight to an \textit{Object Layout} loss defined on each predicted object layout $\hat{l}$ and the corresponding ground-truth layout $\textbf{l}$: $\mathcal{L}_{\text{layout}} = \sum\nolimits_i^n \lVert \textbf{l}_{i} - \hat{l}_{i} \rVert_1$.

\begin{figure}[t]
\begin{center}
\centerline{\includegraphics[width=\linewidth, trim={0 2250 0 0}, clip]{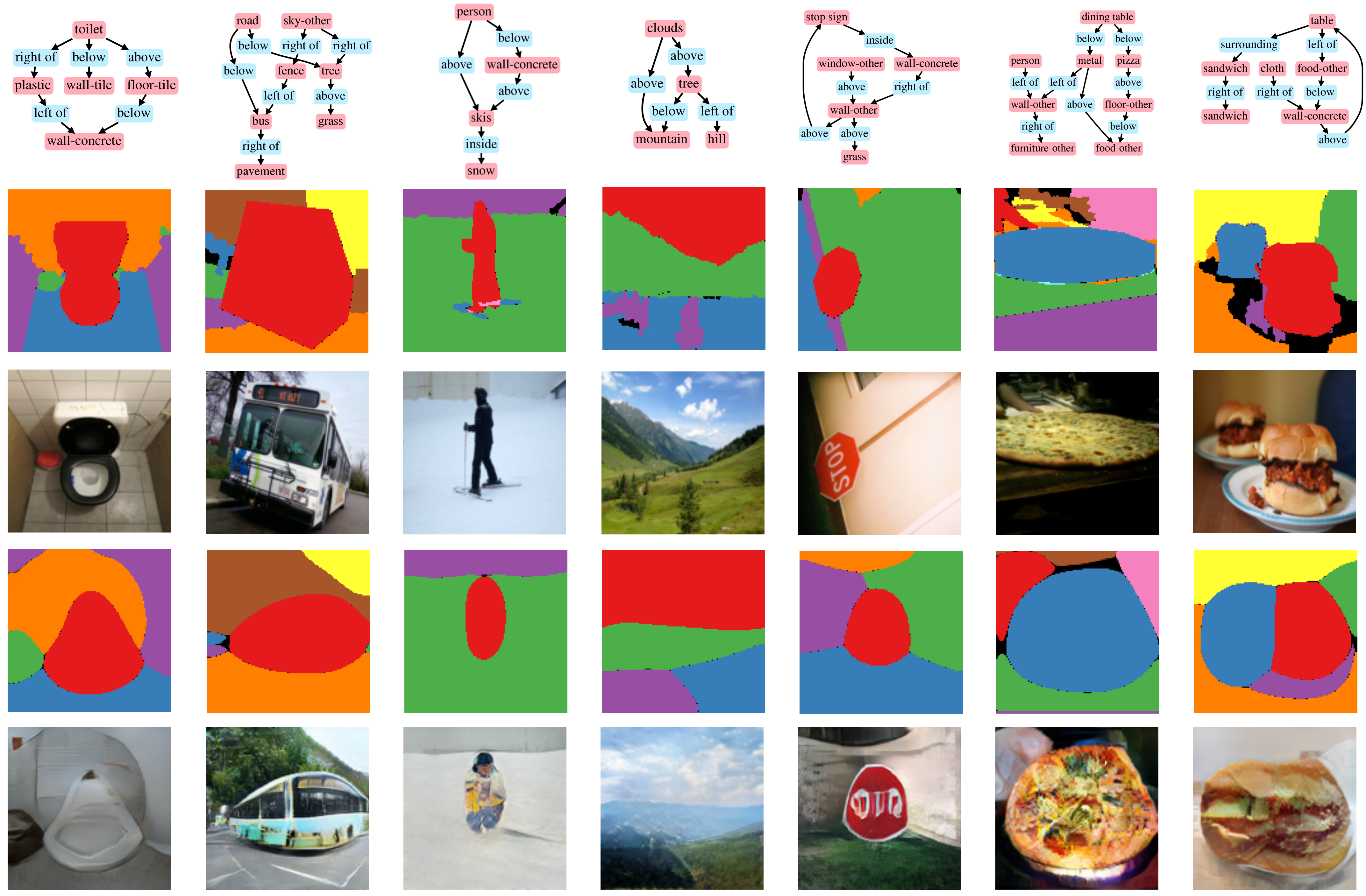}}
\centerline{\includegraphics[width=\linewidth, trim={0 0 0 1700}, clip]{images/color_samples_tighter.png}}
\vskip -0.1in
\caption{Examples of COLoR generated images from SGs.}
\label{fig:color-results}
\end{center}
\vskip -0.4in
\end{figure}

\noindent{\bf Mapping embeddings to layouts directly\quad}
In prior work, boxes $b_i$ and masks $m_i$ were decoded from object embeddings $\tilde{v}_i$ in parallel. Hence, the embeddings $\tilde{v}_i$ computed by the GCN must encode all the information about the location and shape of each object, including avoiding inter-object overlap and maintaining high-coverage of the layout. Furthermore, since the dimension of the mask is $W_m \times W_m$, which is then warped into $H \times W$, and $W_m \ll \min(H,W)$, we can expect a drop in resolution (i.e. very coarse shapes).
To mitigate these issues while remaining agnostic to the SG size and structure, we propose an adaptation of the GCN technique to improve the object layouts generation process by contextualizing object layouts on each other. We name this module \textit{Layout Refinement Network} (LRN).
Formally, given some intermediate object representations $\hat{v}_1^t,\dots,\hat{v}_n^t \in \mathbb{R}^{C_t \times H_t \times W_t}$, we describe a model that predicts the next representations down the line $\hat{v}_1^{t+1},\dots,\hat{v}_n^{t+1} \in \mathbb{R}^{C_{t+1} \times H_{t+1} \times W_{t+1}}$, with $H_{t+1} = 2\cdot H_{t}$, $W_{t+1} = 2\cdot W_{t}$, $C_{t+1} \le C_{t}$.
\begin{equation}
    \hat{v}_1^{t+1},\dots,\hat{v}_n^{t+1} = LRN^t(\hat{v}_1^{t},\dots,\hat{v}_n^{t})
\end{equation}
First, each representation $\hat{v}_{n'}^t, {n'} \in [1,n]$ is passed through a decoder $U$ that applies a transposed convolution layer to upsample the representation by a factor of two, followed by a batch normalization layer and a ReLU activation; $q_{n'}^t = U(\hat{l}_{n'}^t)$.
Each pair of upsampled representations ($\{q_i^t,q_j^t\} | i \ne j \in [1,n]$) is then passed through a graph convolution layer. Due to the dimensionality of the representations ($C_t \times H_t \times W_t$), the traditional dense layers of the graph convolution are replaced with 2D-convolutional layers: $q_{i,j}^t,q_{j,i}^t = GCL^t(q_i^t,q_j^t)$\ .
Summing the initial and pairwise representation produce a new representation that is contextualized on all objects in the scene: $ \hat{v}_{n'}^{t+1} = q_{n'}^t + \frac{1}{n-1} \left ( \sum_{i\ne n} q_{{n'},i}^t + \sum_{i\ne {n'}} q_{i,n}^t \right )$.
In intermediate stages, the residual sum is followed by a ReLU activation. In the final stage, a sigmoid is applied to create the object layout.
Stacking $T$ LRN blocks (where $T=\log_2{H}-1$ due to the behavior of transposed convolutions in the first stage), we skip box and mask predictions Entirely. Instead, our model (depicted in Figure.~\ref{fig:architecture}), generates the layout directly from object embedding. Samples are shown in Figure.~\ref{fig:color-results}. 
Given embeddings $\tilde{v}^1_i$ of size $1 \times 1$ and depth $K=2^T$, we stack $T$ layers of upsampling which reduces the depth by a factor of two followed by an LRN. The output of the model is a set of object layouts $\hat{l}_1,\dots,\hat{l}_n \in [0,1]^{H \times W}$. In each stage, the LRN is used to pass information between the layouts which results in a coherent layout exhibiting extremely high coverage and negligible overlaps. 
We name our model \textit{Contextualized Objects Layout Refiner} (COLoR).
To reduce computational constraints and allow for diversity in the generation, we sample a random subset of all possible layout pairs in each layer.

\begin{table}[t]
\centering
\begin{tabular}{|@{~}l@{~}l@{~}|@{}c@{~}c@{~}c@{~}c@{}|@{}c@{~}c@{}|}
\hline
& & \multicolumn{4}{c|}{Layout} & \multicolumn{2}{c@{}|}{Image} \\
& Model & \small{COV$\uparrow$} & \small{OVL$\downarrow$} & \small{DEC$\uparrow$} & \small{GRS$\uparrow$}  & \small{FID$\downarrow$} & \small{DIV$\uparrow$} \\
\hline
\hline
\parbox[t]{3mm}{\multirow{4}{*}{\rotatebox[origin=c]{90}{Baselines}}}
& SG2Im\cite{Johnson_2018_CVPR}      & 0.76 & 0.18 & 0.83 & 0.62 & 124.3 & 0.0   \\
& SG2Im+SPADE                                   & & \multicolumn{2}{c}{unchanged} & & 97.7 & 0.05\\
& Grid2Im\cite{ashual2019specifying}      & 0.81 & 0.17 & 0.98 & 0.94 & 96.4  & 0.0  \\
& Grid2Im+SPADE                           & & \multicolumn{2}{c}{unchanged} & & 106.6 & 0.0\\
\hline
\parbox[t]{3mm}{\multirow{3}{*}{\rotatebox[origin=c]{90}{Ours}}}
& COLoR                                             & \textbf{0.99} & \textbf{0.0} & \textbf{1.0} & \textbf{0.97} & \textbf{95.8} & 0.13  \\
& \qquad \  - $\mathcal{L}_{D_l}$                   & \textbf{0.99} & \textbf{0.0} & \textbf{1.0} & 0.55 & 102.9 & 0.09  \\
& \qquad \  - $\mathcal{L}_{\text{layout}}$         & 0.82 & \textbf{0.0} & \textbf{1.0} & 0.60 & 122.7 & \textbf{0.50} \\
\hline
\end{tabular}
\vskip -0.1in
\caption{Layout quality evaluation with Coverage, Overlap, Decisiveness, and Geometric-Relation-Score. Image quality evaluation with FID and Perceptual Diversity.}
\label{table:benchmarks}
\vskip -0.2in
\end{table}

\section{Experiments}

We train our model to generate 128x128 layouts and use a pretrained SPADE~\cite{Park_2019_CVPR} model to predict images from them. We show that it creates high-quality layouts, respects the SG's constraints and results in overall higher quality images compared to prior work. We follow the setup in \cite{Johnson_2018_CVPR, ashual2019specifying} using the \cocostuff dataset~\cite{caesar2018coco}.
This dataset contains a subset of the images in COCO~\cite{lin2014microsoft} with additional 91 \textit{stuff} categories. 
We compare our model against a pretrained model of \cite{ashual2019specifying}, and a model of \cite{Johnson_2018_CVPR} trained to produce images of the same resolution. We evaluate layout quality, diversity and adherence to SGs and the predicted image's quality.  

\noindent{\bf Layout generation\quad}
To evaluate the quality of the generated layouts, we measure the average coverage, overlap, and decisiveness of the layouts. Coverage ranges between $0$ to $1$ where higher values are better as it means the layout does not contain empty spots. Overlap measures if multiple objects occupy the same pixel which we wish to avoid. The decisiveness measure evaluates how decisive the generator is in deciding the pixels' pertinence.
Formally, given predicted layouts $\hat{l}_1,\dots,\hat{l}_n \in [0,1]^{H \times W}$ we threshold the layouts $\hat{l}^t_{i,h,w} = \mathbbm{1}[\hat{l}_{i,h,w} \ge t]$ setting $t=0.5$
to get $\hat{l}_1^t,\dots,\hat{l}_n^t \in \{0,1\}^{H \times W}$. We then define the coverage as $\frac{1}{M}\sum_{h,w}{\mathbbm{1}[\sum_i^n{\hat{l}^t_{i,h,w}}\ge 1]}$, the overlap as $\frac{1}{M}\sum_{h,w}{\mathbbm{1}[\sum_i^n{\hat{l}^t_{i,h,w}}\ge 2]}$, and the decisiveness is defined as $\frac{4}{M}\sum_{i,h,w}{ (0.5-\hat{l}_{i,h,w})^2}$. Where $M = H \times W$.

Finally, to evaluate the compliance with relation constraints, we define the \textit{Geometric-Relation-Score} (GRS). Given a pair of predicted object layouts $\hat{l}_i, \hat{l}_j$, we compute the minimal axis-aligned bounding rectangle that contains all pixels with values above 0.5, and projecting it to $W_m \times W_m$ to get a mask. We then use the same heuristic that was used in the construction of the dataset to infer the relations between objects in the predicted layouts, and define the geometric relation score as the accuracy of these predictions. 

\noindent{\bf Image generation\quad}
To evaluate the quality of the generated images, we use the common \textit{FID} score.
We augment this evaluation with the diversity measure suggested by \cite{ashual2019specifying}, which relies on the Perceptual Similarity measure~\cite{zhang2018unreasonable}. There, multiple images are generated from the same SG, and we measures the average distance between every pair,
where large distance is correlated with diversity.

\noindent{\bf Results\quad}
As depicted in Table~\ref{table:benchmarks}, our method outperforms on all \textit{layout generation} benchmarks by large margins. It predicts layouts that have both high coverage and low overlap. In addition, the layouts are decisive and fulfill the geometric relations specified by the SG.
The image generation quality of our model is preferable compared to the baselines according to both the FID measure and the diversity score. Our model achieves an FID and diversity score of 95.8 and 0.13 respectively. 
Our model also shows more diversity when generating multiple images from the same SG than the baselines. It should be noted that one of the ablations (-- $\mathcal{L}_{\text{layout}}$) scored very high on the diversity benchmark due to its failure to generate reasonable layouts consistently, which means that diversity on its own is not sufficient, and should always be evaluated in conjunction with an image quality benchmark.

\noindent{\bf Ablations\quad}
We study the contributions of $\mathcal{L}_{D_l}$ and $\mathcal{L}_{\text{layout}}$ by training COLoR models without them. It can be seen in Table~\ref{table:benchmarks} that removing the pairwise layout discriminator ${D_l}$ impairs the GRS and removing $\mathcal{L}_{\text{layout}}$ hurts the layout coverage.
To show that improvements in the final image quality are due to improved layouts, and not due to SPADE's superiority over prior work's layout-to-image networks, we evaluate the image generation quality of SG2Im~\cite{Johnson_2018_CVPR} and Grid2Im~\cite{ashual2019specifying} layouts by replacing their layout-to-image modules (CRN~\cite{chen2017photographic} and Pix2Pix~\cite{isola2017image} respectively)  with the SPADE~\cite{Park_2019_CVPR} model we use. We find that the FID score is heavily dependent on the type of generator and not necessarily on the quality of the layout, which was the main focus of this work. Both SG2Im and Grid2Im scored differently using their own generators compared to using SPADE. However, SG2Im score improves while Grid2Im suffers. 

\section{Conclusions}
We presented a new technique to train a model to directly predict object layouts from an abstract scene description while attending all objects simultaneously. Our method achieves a sizable improvements in the layouts' quality compared to prior works, resulting in accurate and photo-realistic images.

\section{Acknowledgements}
This project received funding from the European Research Council (ERC) under the European Unions Horizon 2020 research and innovation programme (grant ERC CoG 725974 and grant ERC DELPHI 802800).
This research was partially supported by The Yandex Initiative for Machine Learning.


\bibliographystyle{IEEEbib}
\bibliography{egbib}

\end{document}